\documentclass[letterpaper, 10 pt, conference]{ieeeconf}  

\IEEEoverridecommandlockouts                          

\usepackage{amsmath} 
\usepackage{graphicx}
\usepackage{caption}
\usepackage{subcaption}
\usepackage{multirow}
\usepackage[normalem]{ulem}
\usepackage[dvipsnames]{xcolor}
\usepackage{array}
\usepackage[switch,pagewise]{lineno}
\usepackage{hhline}
\title{\LARGE \bf
Learning to Switch CNNs with Model Agnostic Meta Learning for Fine Precision Visual Servoing
}

\author{Prem Raj$^{1}$, Vinay P. Namboodiri$^{1,2}$ and L. Behera$^{1,3}$ 
\thanks{Affiliations: $^{1}$ Indian Institute of Technology, Kanpur. $^{2}$ University of Bath. $^{3}$ TCS Innovation Lab, Noida. Emails:{\{praj, vinaypn, lbehera\}@iitk.ac.in}}
}

\begin{document}

\maketitle
\thispagestyle{empty}

\pagestyle{empty}

\begin{abstract}
\label{abstract}
Convolutional  Neural  Networks (CNNs)  have been successfully applied for relative  camera  pose estimation from labeled image-pair data, without requiring any hand-engineered features, camera intrinsic parameters or depth information. The trained CNN can be utilized for performing pose based visual servo control (PBVS).
One of the ways to improve the quality of visual servo output is to improve the accuracy of the CNN for estimating the relative pose estimation. With a given state-of-the-art CNN for relative pose regression, how can we achieve an improved performance for visual servo control? In this paper, we explore switching of CNNs  to improve the precision of visual servo control. The idea of switching a CNN is due to the  fact that the dataset for training a relative camera pose regressor for  visual servo control must contain variations in relative pose ranging from a very small scale to eventually a larger scale. We found that, training two different instances of the CNN, one for large-scale-displacements (LSD) and another for small-scale-displacements (SSD) and switching them during the visual servo execution yields better results than training a single CNN with the combined LSD+SSD data. However, it causes extra storage overhead and switching decision is taken by a manually set threshold which may not be optimal for all the scenes. To eliminate these drawbacks, we propose an efficient switching strategy based on model agnostic meta learning (MAML) algorithm. In this, a single model is trained to learn parameters which are simultaneously good for multiple tasks, namely a binary classification for switching decision, a 6DOF pose regression for LSD data and also a 6DOF pose regression for SSD data. The proposed approach performs far better than the naive approach, while storage and run-time overheads are almost negligible.  

\end{abstract}

\section{INTRODUCTION}
\label{sec:intro}
In visual servoing, motion of a robot is controlled by integrating visual feedback into robot control \cite{Chaumette2006VisualSC}. Typically an eye-in-hand configuration is used for visual feedback, in which a camera is mounted on the end-effector of the robot.
The objective is to move the camera from an arbitrary 6DOF camera pose (position+orientation) to a fixed goal pose indicated by the corresponding reference image. The movement of the camera is achieved by iteratively minimizing the error between the current pose and the goal pose.
Classical visual servoing algorithms can be divided into two categories - position-based visual servo (PBVS) \cite{PBVS2} and image-based visual servo (IBVS) \cite{IBVS1}. In PBVS, the error between estimated current camera pose and the given reference pose is defined in the Euclidean space. The bottleneck for PBVS is that it requires to know the camera parameters and the 3D geometry of the scene.
In IBVS, error in image space is minimized using tracking and matching of handcrafted visual features such as points, line and moments. IBVS also requires intrinsic camera calibration to transfer the feature values from image frame to camera frame \cite{Chaumette2006VisualSC}. Choosing the right set of features is very crucial to the IBVS method which is a tedious task. 

An alternative approach called `direct visual servoing' (DVS) \cite{DVS2}, eliminates the need for extracting and tracking the handcrafted features as it uses the image intensities directly in the computation of the interaction matrix. However, it suffers from reduced domain convergence.

Recently, CNN based methods have achieved remarkable success in various robotics tasks such as object grasping \cite{Grasp2}, navigation \cite{Navigation1}, automated bin/shelf picking \cite{picking} etc.      
CNN based methods have also been explored to solve the visual servoing problem as well \cite{Servoing1,Servoing2,servoing3,servoing4}. Similar to DVS approach, CNN based methods also do not require  handcrafted features, depth information, and intrinsic camera calibration. Moreover, CNN based methods have shown greater domain convergence than DVS \cite{Servoing1}.

We undertake the problem of eye-in-hand pose based visual servo control where CNN is trained to estimate relative camera pose. The dataset consists of image-pairs and the corresponding relative camera pose between them as ground truth labels. Better the pose estimation by CNN, better is the visual servo control. How best visual servo control can be performed with a given state-of-the-art CNN for relative pose regression? 
In this paper, we introduce a novel idea of CNN switching to achieve better precision for visual servo control. Our approach is generic and can be applied to problems of similar kind. 
The idea of CNN switching is based on the observation that the dataset to train the relative camera pose regressor for visual servo control, must contain variations in relative pose, ranging from a smaller scale to a larger scale. It is found that adding an extra amount of data generated with a finer sampling (small scale camera displacements) improves the precision of visual servo control at finer scale \cite{Servoing1}. With this inspiration, we generate two different datasets, namely large scale displacements (LSD) dataset, which is generated by sampling camera displacements in a large enough range covering the designated setup area and another is small scale displacements (SSD) dataset, which is generated by sampling camera displacements in a very small range (refer to Tab. \ref{table:data-lims}). 

What is the optimal way to train our model(s) on such combination of data so that it achieves the best performance for visual servo task?
In this paper, we present a novel and unique way to train our CNN model on this combination of data which outperforms naive vanilla method in terms of accuracy of visual servo control and also improves over other comparable methods with fewer parameters. The training method uses model-agnostic-meta-learning (MAML \cite{maml}) algorithm and exploits the idea of CNN switching to achieve finer precision visual servo output.

\section{Related works}
\label{sec:related}

Recently, there have been several works on visual servoing based on CNNs. These works can be categorized in two ways based on how the visual servoing problem is formulated. In one way, the task for the robot is to position itself in the 6DOF cartesian space such that the current camera view matches best with the given reference image \cite{Servoing1,Servoing2}. 
In the other way, the task for the robot is simply to reach near a target object, which is indicated by an image of the object \cite{servoing3,servoing4}; the final orientation of the end-effector can be arbitrary and there is no need for it to match with reference image exactly. 
In this paper, we formulate the visual servoing problem similar to the former way.

In \cite{Servoing1}, two different strategies are presented for visual servoing. In their basic servoing strategy, for each reference image, 10K training samples are randomly sampled around reference image to train CNN. Though visual servo control is able to converge during run time, collecting 10K samples and training a CNN for each new reference image, is impractical. 
In their second strategy, reference image is not assumed to be fixed. Pretrained VGG16 based CNN is trained with 100K training samples. However, it only achieves a broad range convergence. To get finer convergence, it was combined with DVS (direct visual servoing) method. This means significant improvements are needed in this strategy to work as a standalone method for visual servoing. 

In \cite{servoing3}, the goal is to direct the manipulator arm closer to the target object as indicated by the reference image. Recurrent deep network with supervised learning and reinforcement learning is used to learn the arm motion. In \cite{servoing4}, visual servoing is used for navigation task in an indoor environment.

In a recent work \cite{zhuang} for a robotic reaching task in an eye-to-hand setup, a hot-swap strategy is used for swapping the global and local estimation network. The strategy differs from our switching strategy at the implementation level. In their strategy, the local network is trained by zoomed-in images of the dataset used for global network to provide the focused estimation. In our case, with an eye-in-hand setup, focused estimation at the finer level is achieved by providing only the small-scale displacements data to the local network. Also, other than the naive threshold based CNN switching approach, we propose an efficient switching strategy based on model-agnostic-meta-learning \cite{maml} algorithm. The meta learning based switching performs far better than the naive approach. 

In an another work \cite{changing_task_objs}, a task switching strategy is discussed to improve automatic fruit harvesting by a robot. For their first task when the robotic arm is still far from the fruits, a low false positive rate is preferred over the fruit detection accuracy. As soon as the arm reaches near the targeted fruit, the second task objective is activated where the fruit detection accuracy is preferred over having a low false positive rate.

\begin{figure*}[t]
     \centering
     \begin{subfigure}{0.45\textwidth}
         \centering
         \includegraphics[width=\textwidth]{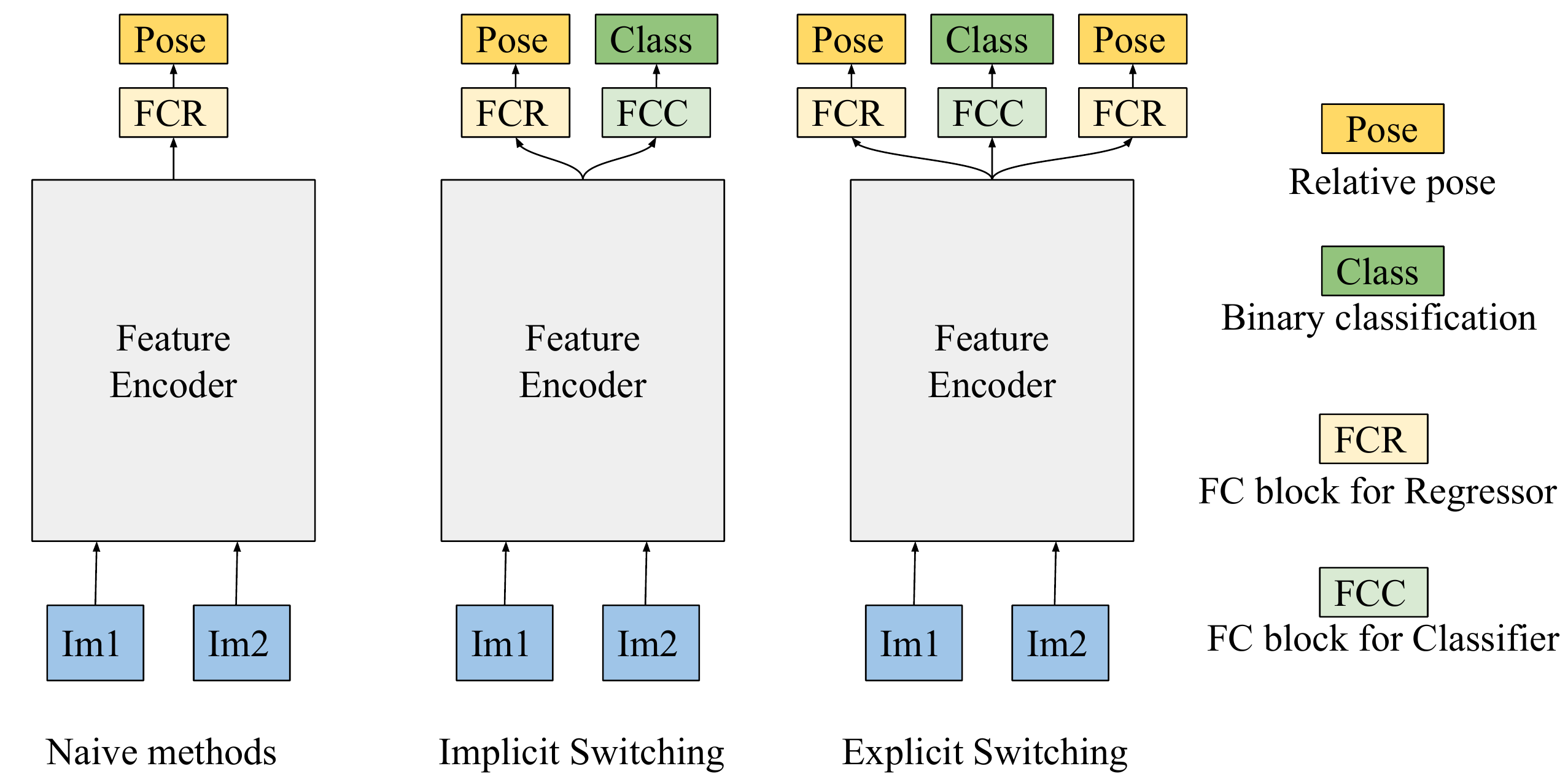}
         \caption{}
         \label{fig:network1}
     \end{subfigure}
     \hfill
     \begin{subfigure}{0.45\textwidth}
         \centering
         \includegraphics[width=\textwidth]{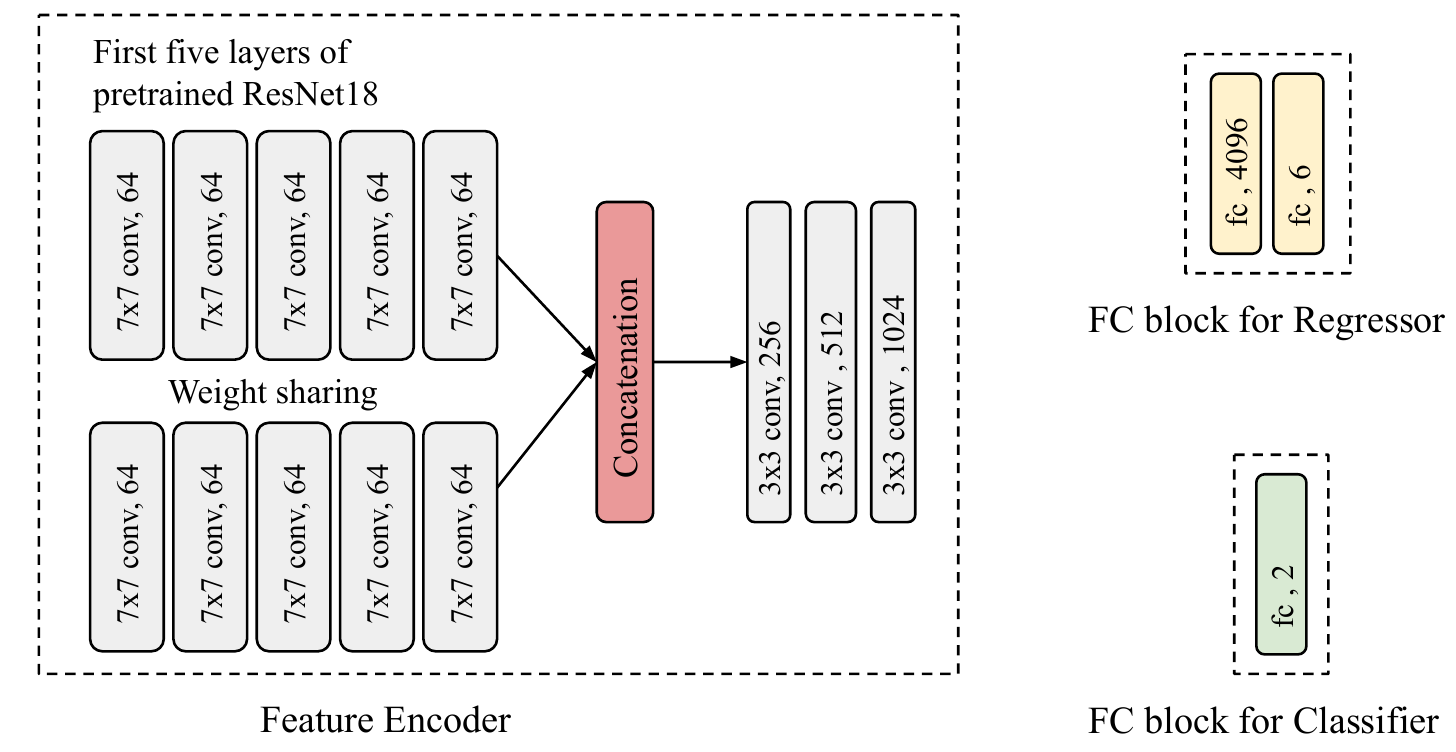}
         \caption{}
         \label{fig:network2}
     \end{subfigure}
     
        \caption{Figure (a) shows three different variants of our pose regression CNN used by different approaches discussed in Sec. \ref{sec:switching}. Figure (b) depicts the different components of our CNN. (Best viewed in color)}
        \label{fig:network}
\end{figure*}

\section{CNN Switching}
\label{sec:switching}
To learn relative camera pose estimation for visual servoing application, the training dataset must contain variations in relative translation and relative rotation ranging from a very small scale to a larger scale.
First, we generate a training dataset named LSD (large scale displacements) as described in Sec. \ref{sec:dataset}. We found that, though, the network trained on this dataset performs good in general for visual servo control, it does not converge to a finer level precision. 
To achieve finer scale convergence, we generate another training dataset namely SSD(small scale displacements) that is having much smaller relative pose ranges compare to LSD (refer to Tab. \ref{table:data-lims}). This time we train our network on the combined data LSD+SSD and found that the precision of visual servo control improves compared to CNN trained only on LSD data.

We propose CNN switching approach to train CNN model(s) which outperforms the above naive methods. Three different variants of this approach are presented in the next subsections.

\subsection{Vanilla Switching}
\label{sec:vanilla-switch}
We found that if we train two separate instances of the CNN with LSD and SSD datasets respectively and at the time of execution of visual servo experiment, if the two instances are switched in their respective domains based on some error threshold then the final visual servo output is much finer than what is achieved with CNN trained with combine LSD+SSD data. 

However, this scheme has two major drawbacks. First, it requires to train and store two instances of the CNN. Secondly, the optimal switching decision is not always trivial. At run time, what we have access to is the reference image of the scene and the current camera view. The switching decision is taken based on the error calculated from these two images. Due to variations in illumination, scene colors etc. any error formulation which is directly based on intensity values or on features which are not invariant to these changes, the error threshold for optimal switching would vary from scene to scene. 
To rectify these drawbacks, we present two other variants of switching method, discussed in the next subsections. 

\subsection{Implicit Switching }
\label{sec:comb_switch}
In this method, switching is not performed explicitly but learn it implicitly during the training through an auxiliary classification task. 
For this, a classification head is added to the feature-encoder along with the existing pose regressor head, as shown in Fig. \ref{fig:network1}. The classification head consists of a single FC layer with just two outputs (i.e. to learn a binary classifier). CNN is trained end-to-end with combined LSD+SSD dataset. While the regressor head learns the relative camera pose regression task, the classification head learns to classify the data between LSD and SSD domains. This way the network is forced to learn features which are discriminative between small scale and large scale camera displacements. Intuitively, it helps network to have its focus on small scale camera displacements which are observed during finer lever convergence of visual servo control.
This simple yet powerful approach outperforms the naive method of training the vanilla CNN with combined SSD+LSD dataset with a greater margin. 


\subsection{Learning switching explicitly through model agnostic meta learning approach}
\label{sec:maml_switch}


For switching to be effective, we need to extract image features such that the error formulation based on these features is invariant to changes such as illumination, scene colors, textures etc. CNN based methods are known to learn features automatically from the data, that are invariant to above changes to a good extent. We learn switching with a binary classifier based on CNN.    
For explicit switching, we learn three different tasks. One is switching which is modeled as a binary classification task to be trained over combined SSD+LSD data and the other two are relative camera pose regression tasks to be trained over LSD and SSD datasets, respectively. We chose a meta learning approach to train our CNN, simultaneously with all three tasks. In our CNN architecture, we add total three heads to the feature-encoder part, one for each task as illustrated in Fig. \ref{fig:network1}. 
 

Meta learning a.k.a. ``learning to learn" aims at training a model on a variety of learning tasks such that it is easily adaptable to a new task. Model agnostic meta learning (MAML) is one such approach in which the weight parameters are learned in such a way that they are simultaneously as good as possible for all the given tasks. Such learned parameters are easily adaptable for any particular task with fewer number of training samples.


Formally, our model is represented by function $f_\theta$ with parameters $\theta$. We want to optimize our model for three different tasks $\{\mathcal{T}_i\}_{i=1,2,3}$. 
In each iteration of MAML, first for each task $\mathcal{T}_i$, with its K training examples (K-shot learning), the parameters $\theta$ are optimized to ${\theta}'_i$ as follows:

\begin{equation}\label{eq:maml1}
     {\theta}'_i = \theta - \alpha \nabla_\theta \mathcal{L}_{\mathcal{T}_i}(f_\theta)
\end{equation}

$\mathcal{L}_{\mathcal{T}_i}$ is the loss function defined for task $\mathcal{T}_i$ and $\alpha$ is the step size hyperparameter.
Model parameters $\theta$ are updated by optimizing the performance of each $f_{\theta'_i}$ obtained in Eq. \eqref{eq:maml1} with the following meta-objective :-

\begin{equation}\label{eq:maml2}
    \min_\theta \sum_{i=1,2,3} \mathcal{L}_{\mathcal{T}_i}(f_{\theta'_i}) = \sum_{i=1,2,3} \mathcal{L}_{\mathcal{T}_i}(f_{\theta - \alpha \nabla_\theta \mathcal{L}_{\mathcal{T}_i}(f_\theta)})
\end{equation}

The aim of the meta-objective is to optimize parameters $\theta$, such that the updated parameters are good for all the tasks. The meta update for the parameters $\theta$ is obtained as follows using simple gradient descent rule :

\begin{equation}\label{eq:maml3}
     \theta \xleftarrow{} \theta - \beta \nabla_\theta \sum_{i=1,2,3} {\mathcal{L}_{\mathcal{T}_i}(f_{\theta'_i})} \exp(-\hat{\mathcal{S}}_i) + \hat{\mathcal{S}}_i
\end{equation}

Where $\beta$ is the meta step size. Term $\exp(-\hat{\mathcal{S}}_i)$ is used to auto balance the losses for all the three tasks, similar to loss function described by Eq. \eqref{eq:loss2}. The balancing weights (i.e. $\exp(-\hat{\mathcal{S}}_i)$) are auto learnt during the training. The losses for different tasks have different ranges and it is important to balance them to get a stable training.

The training examples used in meta update are different than that used for updates of Eq. \eqref{eq:maml1}. We divide the training set of each task into two equal parts, one is used for updates of Eq. \eqref{eq:maml1} and other is used for meta-objective updates of Eq. \eqref{eq:maml3}. 

First, CNN is trained with MAML algorithm described above. Then, each head is finetuned separately, while keeping other parts of the model parameters frozen. This simple yet elegant technique enables to learn all three tasks with just a single model. In Sec. \ref{sec:switch-compare}, we show by empirical results that the MAML training approach gives superior performance as compared to that of all baseline approaches.

\section{Task set-up and model training}
\label{sec:method}

\subsection{Visual servo control law}
\label{sec:overview}
An eye-in-hand configuration is assumed where the camera is attached to the end effector of the manipulator arm. The goal is to position the camera to a target 6DOF pose in the cartesian space such that the current camera view matches the best with the given reference view.
To achieve this our visual servo system predicts the velocity (linear and angular in 3D) of the camera in its local frame, iteratively minimizing the error between the current camera image and the reference image. The estimation is done using pose based visual servo control \cite{Chaumette2006VisualSC} given as follows:-

\begin{equation}\label{eq-control_law}
  \begin{cases}
     \mathbf{v_c} = -\lambda \mathbf{R}^T\;  \mathbf{^{c^*}t_c} \\
     \boldsymbol{\omega_c} = -\lambda\; \boldsymbol{{\theta}_u}
  \end{cases}
\end{equation}

Translation vector $^{c^*}t_c$ gives the coordinates of the origin of the current camera frame ($c$) expressed relative to the desired camera frame ($c^*$).
Matrix $R$ gives the orientation of the current camera frame relative to the desired camera frame and ${\theta}_u$ is the angle parameterization for the rotation matirx $R$.  
Our trained CNN estimates the relative camera pose $(^{c^*}t_c$, ${\theta}_u)$ required by the control law.

\begin{figure}[h]
     \centering
     \begin{subfigure}[b]{0.25\textwidth}
         \centering
         \frame{\includegraphics[width=0.9\linewidth]{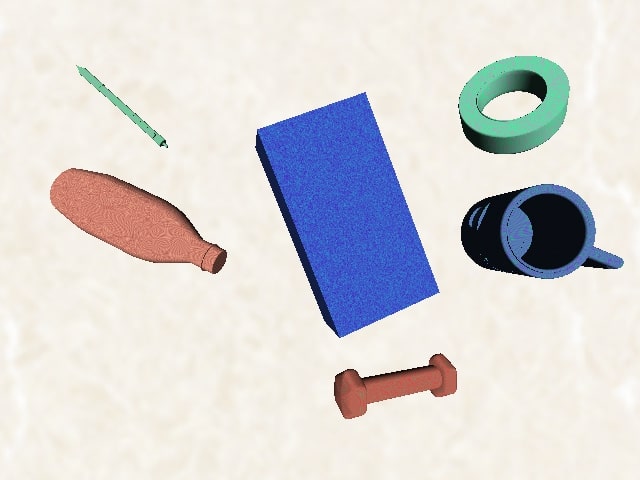}}
         \caption{Simulation}
         \label{fig:data-sim}
     \end{subfigure}%
     \begin{subfigure}[b]{0.25\textwidth}
         \centering
         \frame{\includegraphics[width=0.9\linewidth]{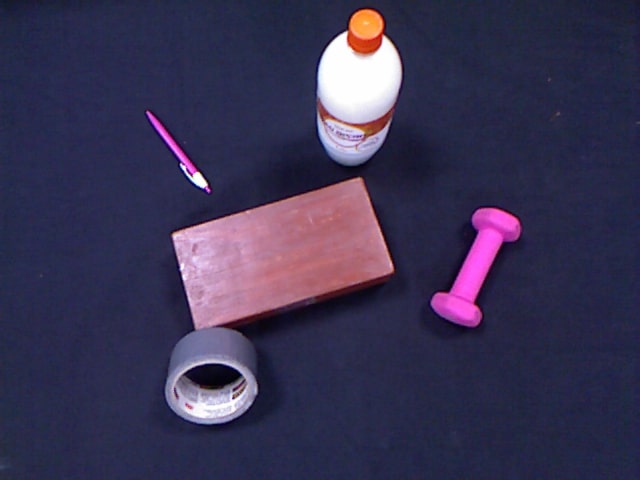}}
         \caption{Real}
         \label{fig:data-real}
     \end{subfigure}%
        \caption{\small Dataset environments}
        \label{fig:data}
\end{figure}

\subsection{Environment set-up and training dataset generation}
\label{sec:dataset}
Training CNN requires enormous amount of data which is difficult to get from real robots. The data generation process for real robots is often manual and time-consuming due to safety concerns and hardware limitations. In \cite{Grasp2}, to generate 50k data samples for a single task, it took 700 robot hours. 
To overcome this difficulty, we create a simulated environment with a target object (i.e. brick) and some distractor objects (i.e. mug, bottle, dumbbell, etc.) as shown in Fig. \ref{fig:data-sim}.

In simulation, we assume free-flying camera model similar to \cite{Servoing2}. Initially, camera is put right above the target object (i.e. brick) to a particular height so that scene is visible nicely. Camera is set to move with random translation and rotation within the predefined limits. For each data sample entire simulation rendering is randomized as mentioned in \cite{DR}. Images at former and later pose of the camera are saved along with the relative pose label. Thus, automatically a large number of data samples are collected effortlessly. 

For real world experiments, we finetune the CNN trained in the simulated environment with few real real world samples (i.e. 100). The real-world task employs an UR10 robotic arm (6DOF) with an eye-in-hand camera setup. Real data samples were collected by manually controlling the robotic arm.

As discussed in Sec. \ref{sec:intro}, we generate two different datasets, namely large-scale-displacements (LSD) dataset and small-scale-displacements (SSD) dataset. 
Camera offset limits for LSD and SSD datasets are given in Tab. \ref{table:data-lims}. The data sampling is done using a equal combination of uniform and gaussian sampling. For each data sampled from gaussian distribution, mean is taken to be zero and standard deviation is chosen randomly between zero and one-third of the upper limit. 


\begin{table}[h]
\centering
\caption{Camera offset limits for data generation. Translations are given in meter and rotations are given in radian.}
\label{table:data-lims}
\begin{tabular}{|c|c|c|} 
\hline
 & LSD & SSD \\ 
\hline
X,Y translation & $[-0.30, 0.30]$ & $[-0.05, 0.05]$    \\
\hline
Z translation & $[-0.20, 0.20]$ &  $[-0.04, 0.04]$   \\
\hline
X,Y rotation & $[-0.15, 0.15]$ &  $[-0.05, 0.05]$   \\
\hline
Z rotation &  $[-0.40, 0.40]$ & $[-0.10, 0.10]$  \\
\hline
\end{tabular}          
\end{table}




\subsection{Network Architecture}
\label{sec:architecture}

In Fig. \ref{fig:network2}, different components of our CNN architecture are depicted which are used for building three different variants as illustrated in Fig. \ref{fig:network1}. Different switching methods described in Sec. \ref{sec:switching} uses one of these variants.
The most significant component of our CNN architecture is the feature-encoder part.
On the top of the feature-encoder, are first five layers of ResNet18 \cite{resnet}, pretrained on ImageNet \cite{imagenet} classification task. Though, originally these layers are trained for a classification task, they are used for relative pose regression task. The top initial layers of any trained CNN are known to produce generic features that can be used for learning a new task \cite{Servoing1}. 
Both, reference image and current image (each of size 224x224) are passed through these top five layers, separately. Subsequently, the output features at the fifth layer are concatenated depth-wise and passed through three more convolutional (conv.) layers to learn features specific to the task. Each of these three conv. layers uses stride 1 and is followed by a `Batchnorm', a `Relu' and a `Pooling' layer, which are not shown in the figure to avoid clutter. At this point, the resultant feature map of size 1024x7x7 is passed through an `AdaptiveAvgPool' layer to produce a feature vector of size 1x1024.



\subsection{Basic training configurations and loss functions}
\label{sec:training}
For all the variants used in this paper, a common training configuration is followed which is found to be effective.
`Adam' optimizer is used with learning rate $10^{-4}$ and weight decay $4\times10^{-5}$ for 50 epoch. The best network from this training is further trained with a slower learning rate of $10^{-5}$ for another 20 epoch which results in improved accuracy. Input images are normalized using per channel mean and variance calculated over the entire training dataset. The use of batch-normalization in intermediate layers of CNN proved useful in terms of increased accuracy. 
For the training, 2 NVIDIA GTX 1080 Ti GPUs is used. 
The loss function is calculated in terms of euclidean distances between predicted and ground truth vectors, for translation(meters) and rotation(radians) respectively as follow, similar to \cite{posenet} - 
\begin{equation}\label{eq:loss1}
     loss(\mathbf{I},\mathbf{I^*}) = \|\mathbf{{\hat{t}}-\mathbf{t}}\|_2 + \beta  \|{\boldsymbol{\hat{{\theta}_u}} - \boldsymbol{{\theta}_u} }\|_2
\end{equation}
Here $\beta$ is used to balance the translation and rotational errors. For the training in our case, $\beta = 0.2$ turns out to be optimal. 

Alternatively, following loss function, proposed in \cite{posenet2}, is used to auto learn the balance weights for different loss terms during the training:
\begin{equation}\label{eq:loss2}
    \mathcal{L} = \sum_{i} \mathcal{L}_i \exp(-\hat{\mathcal{S}}_i) + \hat{\mathcal{S}}_i
\end{equation}
The term $\exp(-\hat{\mathcal{S}}_i)$ represents the homoscedastic uncertainty corresponding to loss term $\mathcal{L}_i$. Additive term $\hat{\mathcal{S}}_i$ acts as self regularizer. 
Balance weights (i.e. $\exp(-\hat{\mathcal{S}}_i)$)  are auto learned in contrast to the loss function given in Eq. \eqref{eq:loss1}, avoiding the difficult task of hyper-parameter tuning. 
However, the final performance of CNN is found to be almost equal in both the cases when either of the loss function is used.  




 \begin{figure}[t]
      \centering
      \includegraphics[scale=0.3]{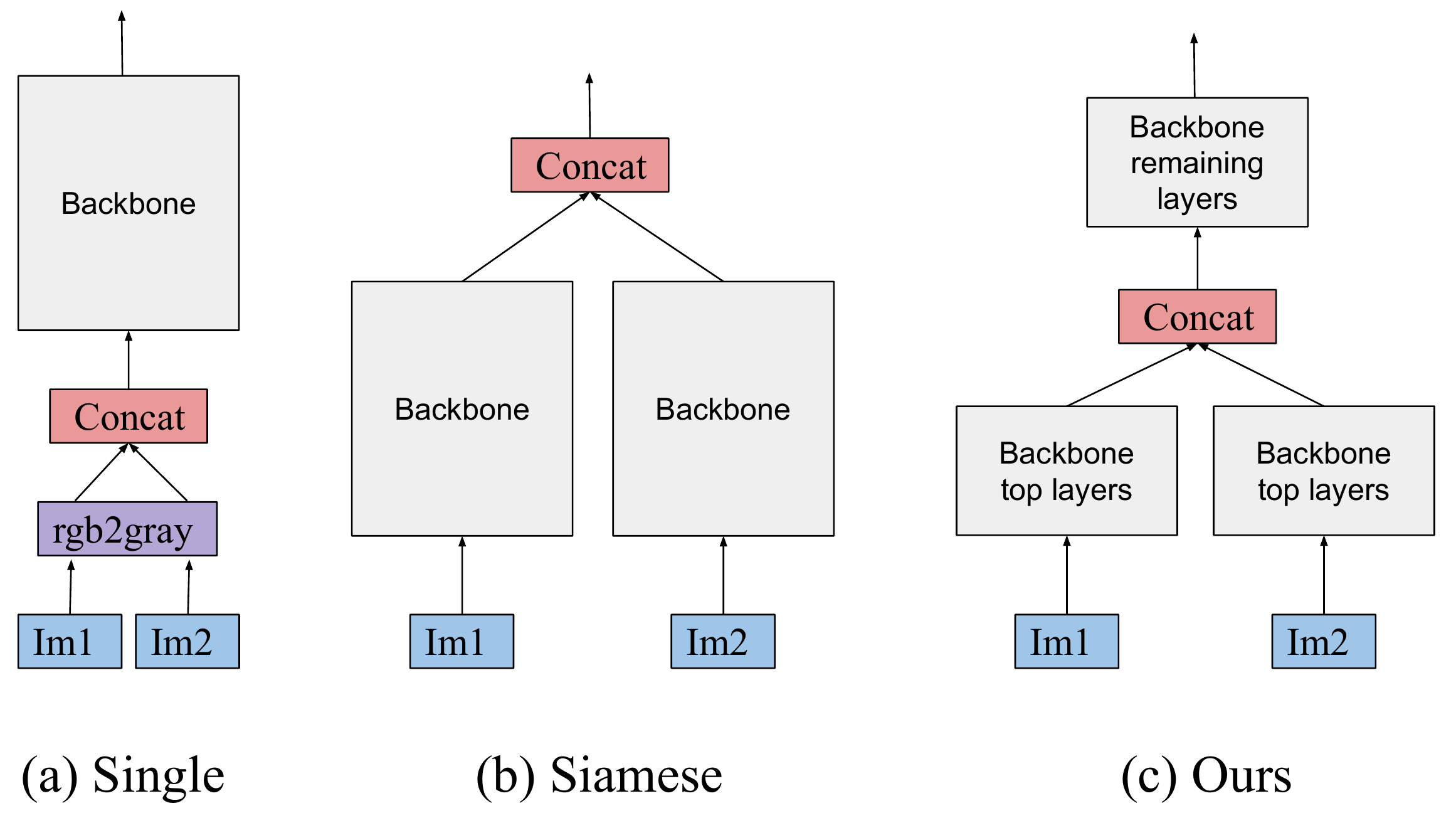}
      \caption{Design choices for feature encoders used in relative pose regressor CNNs, with a given backbone network.
      }
      \label{fig:cnn_compare}
 \end{figure}

\section{Evaluations and Discussions}
\label{sec:results}
We conduct both, simulated evaluations and real-world evaluations.
Simulated evaluations are performed, using Mujoco \cite{mujoco}, with free-flying camera model. Real-world evaluations are carried out using UR10 arm manipulator, with a calibrated eye-in-hand camera. For all the experiments, scenes are static and consist of 3D objects placed on a table.



\begin{figure*}[h]
     \centering
     \begin{subfigure}[b]{0.25\textwidth}
         \centering
         \frame{\includegraphics[width=0.9\linewidth]{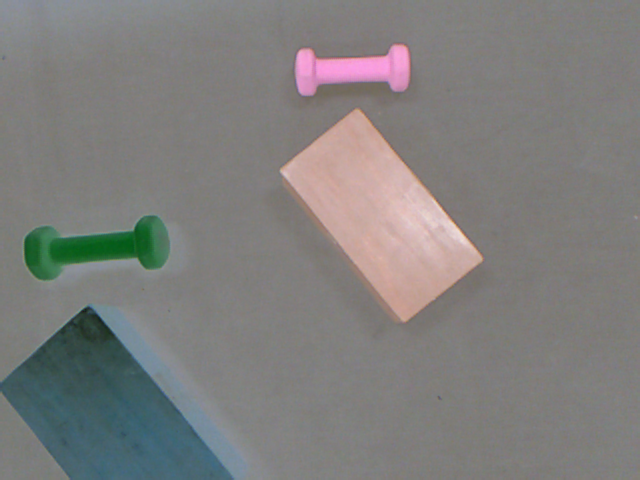}}
         \caption{Start Error Image}
         \label{fig:switch-a}
     \end{subfigure}%
     \begin{subfigure}[b]{0.25\textwidth}
         \centering
         \frame{\includegraphics[width=0.9\linewidth]{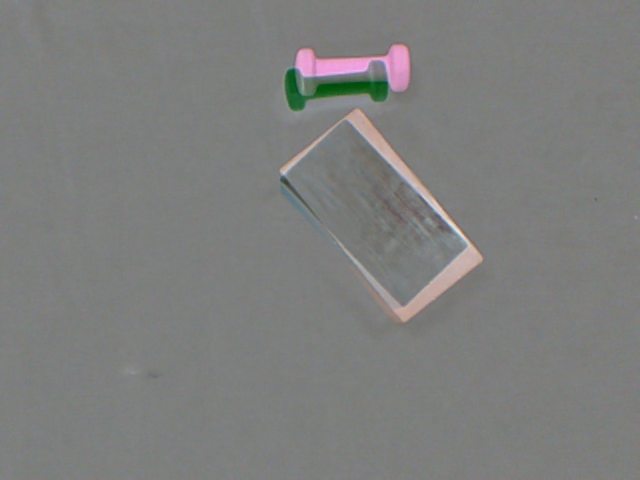}}
         \caption{`LSD only'}
         \label{fig:switch-b}
     \end{subfigure}%
     \begin{subfigure}[b]{0.25\textwidth}
         \centering
         \frame{\includegraphics[width=0.9\linewidth]{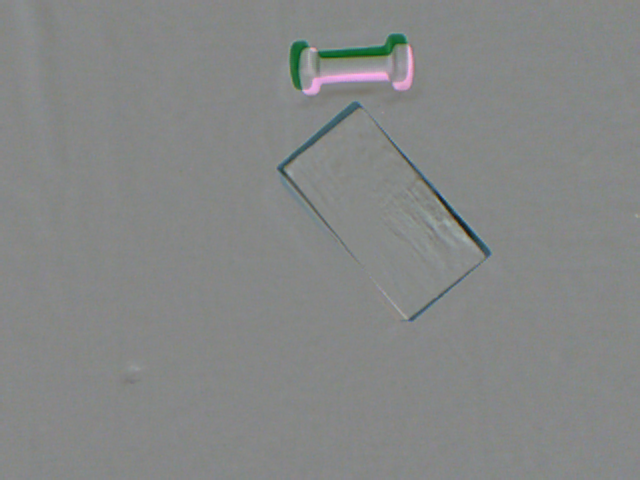}}
         \caption{`Comb'}
         \label{fig:switch-c}
     \end{subfigure}%
     \hfill
     \begin{subfigure}[b]{0.25\textwidth}
         \centering
         \frame{\includegraphics[width=0.9\linewidth]{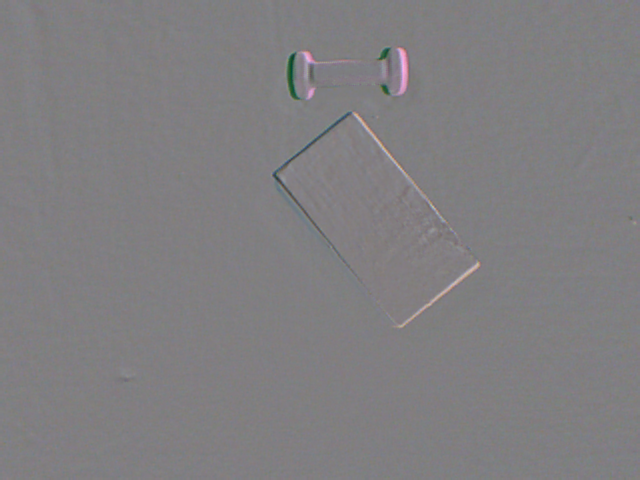}}
         \caption{`Switch(LSD/SSD)'}
         \label{fig:switch-d}
     \end{subfigure}
     
     \begin{subfigure}[b]{0.5\textwidth}
         \centering
         \includegraphics[width=0.9\linewidth]{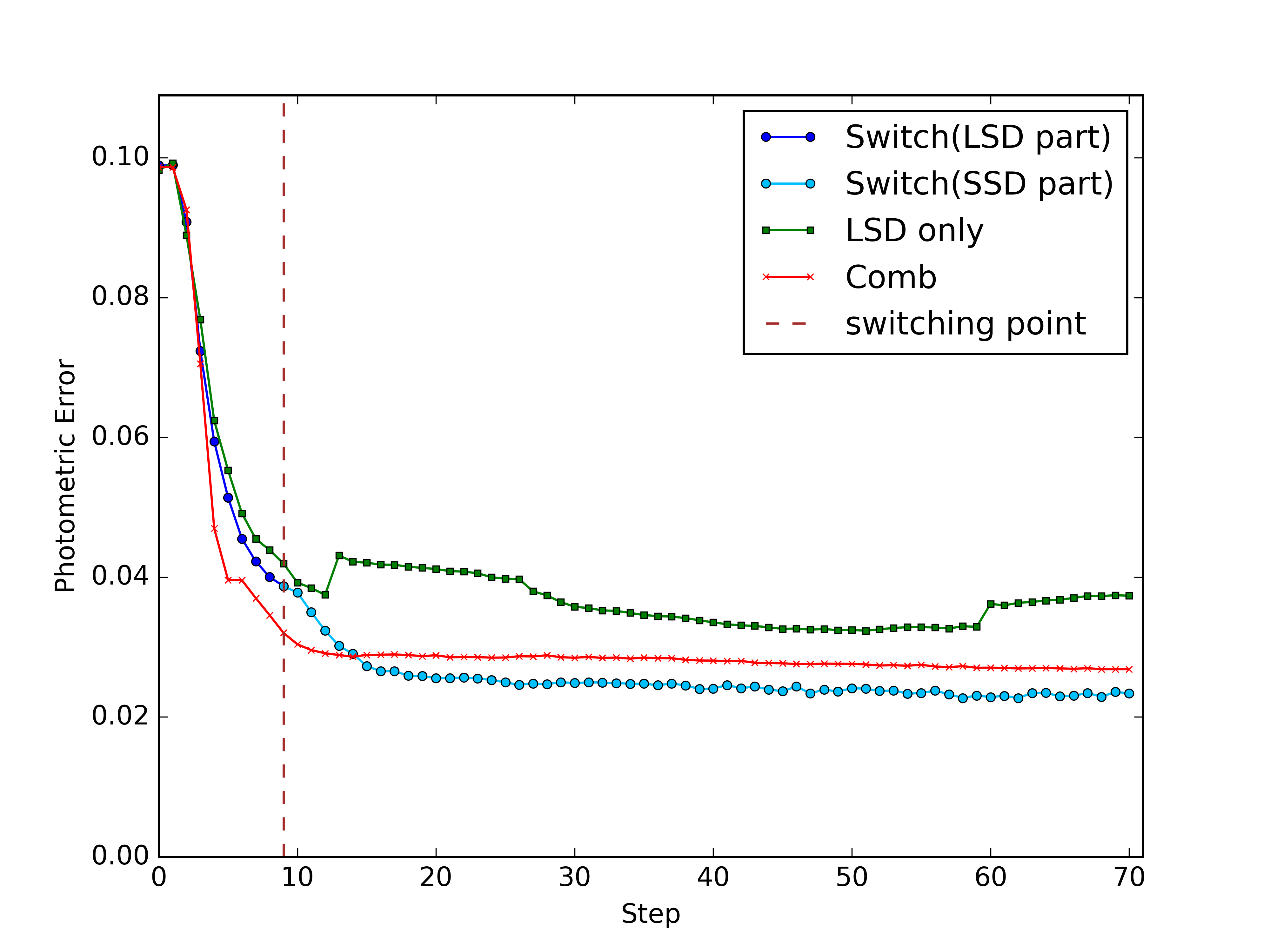}
         \caption{Photometric Error}
         \label{fig:switch-e}
     \end{subfigure}%
     \begin{subfigure}[b]{0.5\textwidth}
         \centering
         \includegraphics[width=0.9\linewidth]{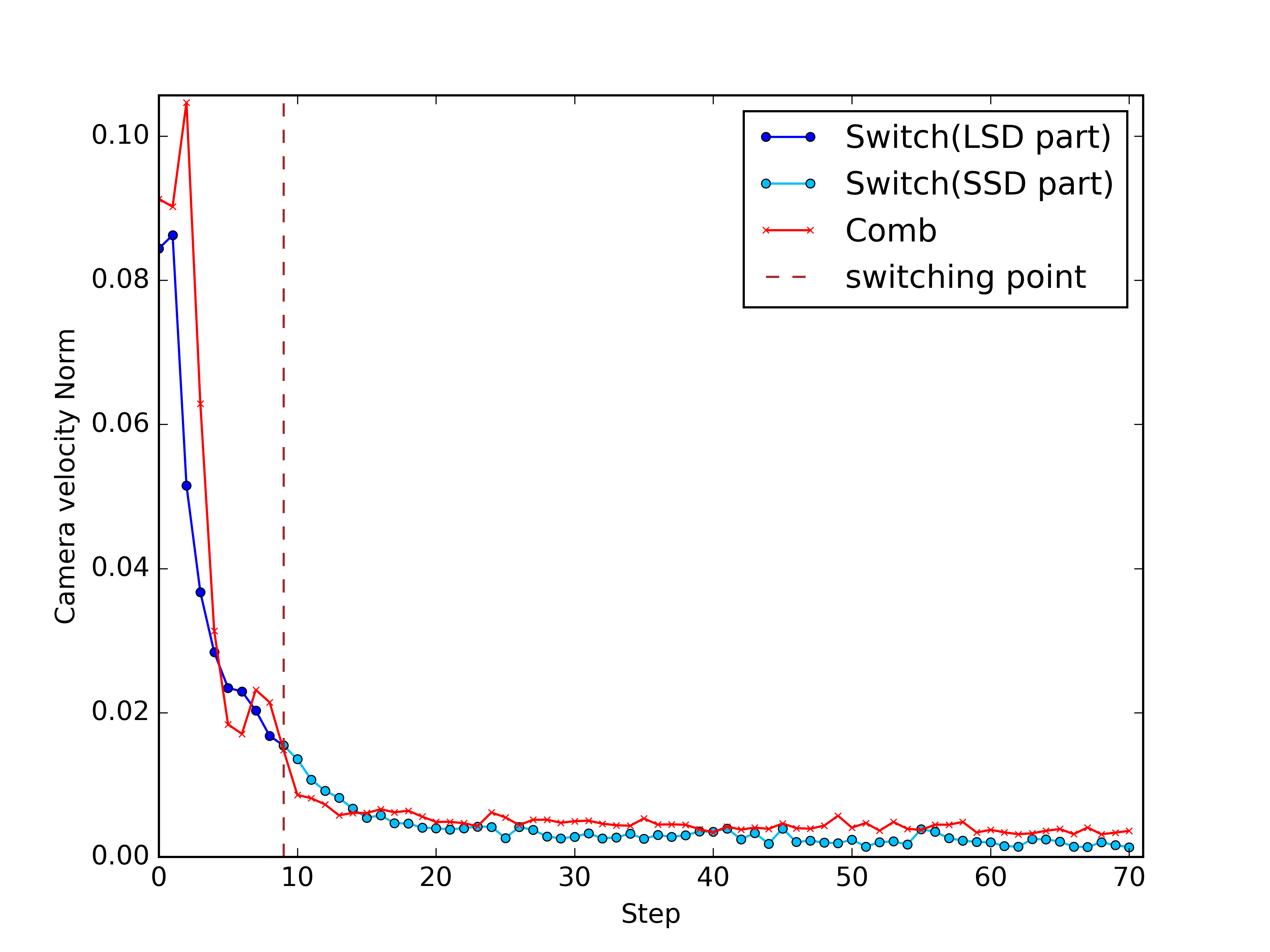}
         \caption{Velocity Norm}
         \label{fig:switch-f}
     \end{subfigure}
        \caption{\small Real-world experiment with UR10 arm to compare the performance of our switching scheme. The experiment runs for 70 steps. (a) Start error Image (b) Final error image by `LSD only' method (c) Final error image by `Comb' method (d) Final error image by our `Switch' method (e) Photometric error plot (f) Camera velocity norm plot.  (Best viewed in color)}
        \label{fig:switch}
\end{figure*}

\subsection{Real world experiment set-up}
\label{sec:ablation-real}
For the real world experiments, the task environment is setup with objects similar to those used for simulations (Fig. \ref{fig:data}). Employing domain randomization \cite{DR} in our simulated dataset, our trained CNN models easily get transferred to real world with simple transfer learning. Total list of attributes for domain randomization and the randomization process is same as given in \cite{DR}. We are able to achieve good performance in real world with finetuning of only FC layers with just a few real world samples (i.e. 100). Domain randomization helps in the sim2real transfer of the CNN learning. To verify that we do an ablation study. Our full method is compared with three variants. Each variant is trained on a dataset generated by randomizing all the attributes but one (the one mentioned in the first column of Tab.~\ref{table:ablation}) and then finetuned with real world data similar to our full method. Average translation and rotation errors for different variants is presented in Tab. \ref{table:ablation}, calculated over 10 real world experiments. Camera offset limits for the experiments are as given in Tab. \ref{table:exp-lims}. It is evident from the results that every attribute contributed in the sim2real transfer of the CNN learning, texture randomization being the most significant. 
Advanced sim2real transfer learning techniques have been proposed recently \cite{DA3,DA2,DA_survey}, however further discussion over them is out of the scope for the current study.

\begin{table}[h]
\centering
\caption{Camera offset limits for quantitative experiments done in Sec. \ref{sec:results}. Translations are given in meter and rotations are given in radian.}
\label{table:exp-lims}
\begin{tabular}{|c|c|c|c|} 
\hline
 & \multicolumn{2}{|c|}{Simulation} & \multirow{2}{*}{Real-world} \\ 
\hhline{~--~}
& proximal & distal &  \\

\hline
X,Y translation  & $[-0.15, 0.15]$ & $[-0.30, 0.30]$ &  $[-0.20, 0.20]$    \\
\hline
Z translation  & $[-0.10, 0.10]$  & $[-0.20, 0.20]$ &  $[-0.15, 0.15]$   \\
\hline
X,Y rotation & $[-0.07 ,0.07]$  & $[-0.15, 0.15]$ &  $[-0.08, 0.08]$   \\
\hline
Z rotation  & $[-0.15, 0.15]$ &  $[-0.40, 0.40]$ &  $[-0.15, 0.15]$  \\
\hline
\end{tabular}          
\end{table}

\begin{table}[t]
\centering
\caption{Ablation study on domain randomization: average results over 10 real world experiments (Sec. \ref{sec:ablation-real})}
\label{table:ablation}
\begin{tabular}{|l|l|l|} 
\hline
Model Variant  & Pos. error (meter) & Rot. error (radian) \\ 
\hline
Full method    & $\bf{0.011} \pm \bf{0.006}$       & $\bf{0.051} \pm \bf{0.030}$        \\ 
\hline
No Texture     & $0.044 \pm 0.034$      & $0.484 \pm 0.073$         \\ 
\hline
No distractor  & $0.035 \pm  0.028$     & $0.118 \pm 0.059$         \\ 
\hline
No light rand. & $0.018 \pm 0.014$       & $0.107 \pm 0.083$           \\
\hline
\end{tabular}
\end{table}

\subsection{Network design comparisons}
\label{sec:cnn_compare}

Here we provide empirical results to assess performance of our CNN compared with previous related works, \cite{Servoing1} and \cite{rpnet}.
These models differ with ours in two aspects, firstly in the backbone network used in feature encoder and secondly in the design choice for feature encoder. Fig. \ref{fig:cnn_compare} depicts three design choices for feature encoder. 
\cite{Servoing1} uses VGG16 \cite{vgg} as the backbone and its feature encoder design is as given in Fig. \ref{fig:cnn_compare}(a), named `single'.
\cite{rpnet} uses GoogleNet \cite{googlenet} as the backbone and a siamese style feature encoder (Fig. \ref{fig:cnn_compare}(b)).
Our feature encoder is depicted in Fig. \ref{fig:network}. Its design corresponds to \ref{fig:cnn_compare}(c).
With three design choices \{single, siamese, ours\} and three backbones \{vgg, googlenet, ours\}, 8 different variants are trained on our LSD dataset. Tab. \ref{table:cnn_design} presents the regression loss obtained by evaluating each variant on our testset. In the table, each variant is named depending upon its design choice and backbone. For example, `siamese-vgg' model uses vgg as the backbone and design choice is `siamese'.

From the results in the table, it is observed that ours design choice of feature encoder has achieved better accuracy compare to `single' and `siamese'. Our model with vgg backbone (i.e. ours-vgg) has achieved slightly lower regression loss than our proposed model (i.e. ours-ours). However ours-vgg has almost 13 times more number of weight parameters, that is 136.4 million (M) compare to 10.5 M parameters in ours-ours model.
Overall, our proposed model (i.e. ours-ours) has achieved relatively better regression loss (i.e. 0.000665) with lesser number of parameters. 
Last column of the table shows the regression loss achieved by our ours-ours method (i.e. 0.000782) when it was trained on auto-balancing loss function described in Eq. \eqref{eq:loss2}.

\begin{table}[h]
\centering
\caption{Comparison: regression loss over test data for different CNN architectures (RL: Regression Loss)}
\label{table:cnn_design}
\begin{tabular}{|l|l|l|} 
\hline
CNN Architecture & RLs & \#Weight parameters  \\ 
\hline
\hline
single-vgg \cite{Servoing1}  & 0.000634  & 134.3 M     \\ 
\hline
siamese-vgg  & 0.000906  & 134.5 M     \\ 
\hline
ours-vgg   & \textbf{0.000606}  & 136.4 M     \\ 
\hline
\hline
siamese-googlenet \cite{rpnet}  & 0.003394  & 7.29 M     \\ 
\hline
ours-googlenet    & \textbf{0.002242}   & 7.12 M    \\
\hline
\hline
single-ours  & 0.001247  & 10.4 M     \\ 
\hline
siamese-ours  & 0.001435  & 14.4 M     \\ 
\hline
\bf{ours-ours (Proposed)} & \bf{0.000665}  &  \bf{10.5 M} \\
\hline
\hline
ours-ours (Loss auto balance) & 0.000782  &  10.5 M \\
\hline
\end{tabular}          
\end{table}

\subsection{Switching benefits over naive approaches (with real-world experiments)}
\label{sec:switch-benefits}
In this subsection, we discuss the benefits of CNN switching in general over the naive approaches with real world experiments. In the subsequent subsection, specific switching algorithms proposed in the paper are evaluated.
To discuss results for this subsection, we use certain shorthand names for different models as follow:\\
\textbf{Switch} :- This method corresponds to the `Vanilla-switch' method discussed in Sec. \ref{sec:vanilla-switch}. Two different instances of CNN are trained, respectively with LSD and SSD dataset. Switching decision is taken based on an error threshold defined over MSE error. \\
\textbf{Switch (LSD part)}:- Part of the experiment carried out with Switch method, where CNN trained with LSD dataset is activated \\
\textbf{Switch (SSD part)}:- Part of the experiment carried out with Switch method, where CNN trained with SSD dataset is activated \\
\textbf{LSD only}:- Method uses only one CNN, trained with LSD dataset \\
\textbf{Comb}:- Method uses only one CNN, trained with combine SSD+LSD dataset \\

\begin{figure}[t]
      \centering
      \includegraphics[scale=0.45]{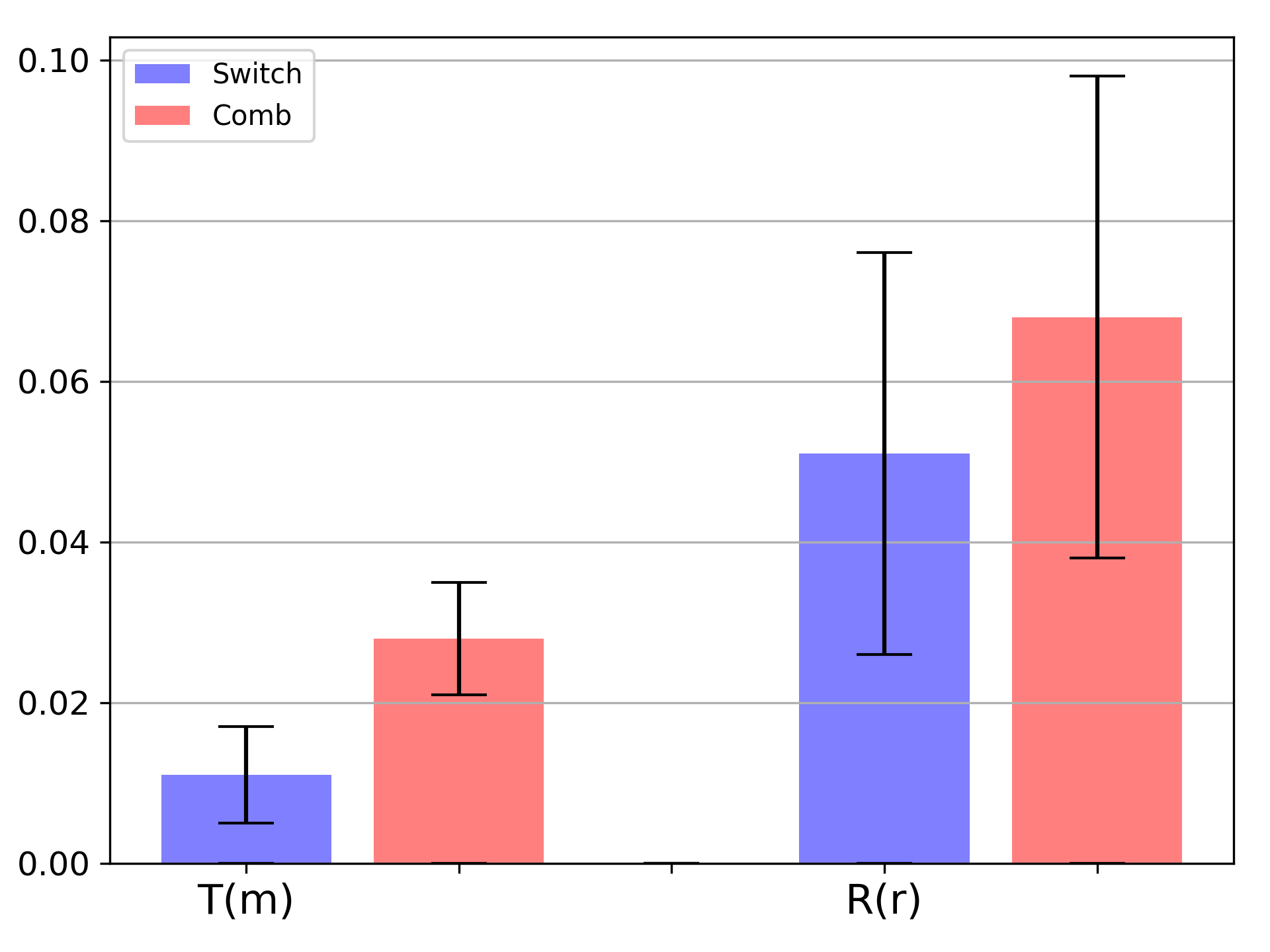}
      \caption{Figure plots the average results obtained over 10 real-world experiments to compare the Switch and Comb methods (refer to Sec. \ref{sec:switch-benefits}).
      T(m):average camera position error in meter; R(r): average camera rotation error in radian
      }
      \label{fig:switch1}
 \end{figure}

In Fig. \ref{fig:switch}, results of a real-world experiment are presented to compare the precision of visual servo output for the above-mentioned methods. In the Figure, start error image, final error images for different methods, photometric error plots, and camera velocity norm plots are given. Further, for quantitative assessment, average position and average rotation errors are plotted in Fig. \ref{fig:switch1}, for 10 real-world experiments. Camera offsets are chosen between the limits given in Tab. \ref{table:exp-lims}.

It is clear, from the results that `Comb' method achieves better precision output than `LSD only' method, (See Figs. \ref{fig:switch-b}, \ref{fig:switch-c} and \ref{fig:switch-e}). This is intuitive as `Comb' method was shown small-scale displacement (SSD) data during its CNN training while `LSD only' was not shown. 
However, `Switch' method achieves even better precision than `Comb' method (See Figs. \ref{fig:switch-c}, \ref{fig:switch-d}, \ref{fig:switch-e} and \ref{fig:switch1}), which indicates the effectiveness of switching scheme in improving precision of visual servo output. The intuition behind the results is that, when CNN is trained with combined data, the estimation of camera pose is kind of averaged out while learning the large scale and short scale camera pose variations simultaneously. On the other side, when CNN instances are trained separately for large scale and short scale datasets, the focused estimation in the respective domains helps achieve better accuracy.

\begin{table*}[h]
\begin{center}
\caption{Comparing different switching strategies, based on average camera position and rotation errors, taken over 100 simulated experiments. The definitions of model names and other details are given in Sec. \ref{sec:switch-compare}. \\(\textbf{PE}: positional error  , \textbf{RE}: rotational error)} 
\label{table:switch2}
\begin{tabular}{l|r|c c||c c} 
  &  & \multicolumn{2}{c||}{proximal} & \multicolumn{2}{c}{distal} \\
\hline
Model Variant & Model size & PE & RE & PE & RE \\ 
\hline
\hline
LSD only & 64.65 MB   & $0.042 \pm 0.059$      & $0.083 \pm 0.133$  & $0.079 \pm 0.339$      & $0.121 \pm 0.409$  \\ 

Comb & 64.65 MB & $0.024 \pm 0.023$     & $0.040 \pm 0.038$  & $0.088 \pm 0.485$     & $0.121 \pm 0.534$\\ 

Vanilla-switch  & 129.30 MB  & $0.039 \pm 0.055$      & $0.073 \pm 0.121$  & $0.130 \pm 0.365$      & $0.154 \pm 0.408$\\ 

CNN-switch & 177.46 MB   & $0.023 \pm 0.021$      & $0.039 \pm 0.034$  & $0.031 \pm 0.083$     & $0.053 \pm 0.116$\\ 

Implicit-switch  & 64.66 MB   & $0.026 \pm 0.023$      & $0.044 \pm 0.038$  & $0.023 \pm 0.023$      & $0.042 \pm 0.038$\\ 
Meta-switch  & 81.16 MB  & $\bf{0.021} \pm 0.020$      & $\bf{0.035} \pm 0.034$  & $\bf{0.023} \pm 0.020$      & $\bf{0.038} \pm 0.034$\\ 
\hline

\end{tabular}
\end{center}
\end{table*}

One question, which may arise whether the switching scheme incurs artefacts in visual servo control either in image space or in Cartesian space? In order to answer this, the consider photometric error plot and the camera velocity-norm plot (Figs. \ref{fig:switch-e} and \ref{fig:switch-f}, respectively). For this particular experiment, switching from LSD to SSD takes place at step `9'. It can be observed from the figures that control remains smooth both in image space (Fig. \ref{fig:switch-e}) and Cartesian space (Fig. \ref{fig:switch-f}).



\subsection{Evaluations of proposed switching schemes (with simulated experiments)}
\label{sec:switch-compare}
In this section, some additional shorthand names are used, which are defined as follows:\\
\textbf{Vanilla-switch}: Two different instances of CNN are trained, respectively for LSD and SSD datasets. Switching decision is based on MSE error threshold (Sec. \ref{sec:vanilla-switch}). \\
\textbf{CNN-switch}: Three different instances of the CNN are trained, one binary classifier for switching purpose and other two for the regression tasks, for LSD and SSD datasets respectively \\
\textbf{Implicit-switch}: A single CNN is trained with combined LSD and SSD data along with an auxiliary classification loss (Sec. \ref{sec:comb_switch}).\\
\textbf{Meta-switch}: A single CNN is trained for three different tasks with model agnostic meta learning approach (MAML) (Sec. \ref{sec:maml_switch}).\\

We evaluate different variants of CNN switching only with simulated experiments. Simulations are a convenient way to perform large number of batch experiments with automated scripts without the fear of safety concerns and hardware failures.

For the experiments in this section, two different scenarios are considered in our simulation environment, namely proximal and distal. Both scenarios have different camera offset limits which are mentioned in Tab. \ref{table:exp-lims}. For experiments in proximal scenario, camera offsets are taken from a small range. In distal scenario, camera offsets are taken from a larger range, thus it exhibits increased difficulty level than the proximal case. For each method in each of the scenario, 100 visual servo experiments are performed. Each single experiment is run for 200 steps of visual servo control and final error values are recorded. The average translation and average rotation errors are calculated and depicted in Tab. \ref{table:switch2}. On the basis of obtained results, the following analysis is done.

CNN-switch performs way better than Vanilla-switch in both proximal and distal cases. This is because, Vanilla-switch takes switching decision based on a fixed error threshold defined over MSE error which is not optimal for all the scenes due to variations in illuminations, colors etc. On the other hand, CNN-switch uses CNN for switch decision which is robust under above variations.

Comb method performs good in proximal case but worse in distal case. This is because, in Comb method, CNN is trained on combined SSD+LSD data and pose estimation is kind of averaged out due to different pose variations in SSD and LSD data. Due to which, Comb method does not perform well, both on very large camera displacements and also on very small scale camera displacements. 
Whereas, Implicit-switch, which is also trained on combined LSD+SSD data, performs well both in proximal and distal cases. This is because Implicit-switch trains an extra auxiliary classification head in CNN to learn discriminative features for SSD and LSD datasets, thus avoiding the averaging out effect which occurs with Comb method.

Meta-switch performs slightly better then Implicit-switch, but it has slightly bigger model size. Meta-switch performs switching explicitly having more focused pose estimation in the respective domains, namely SSD and LSD.  

Although, CNN-switch too performs switching explicitly, it also trains three different instances of the CNN, namely for switching decision, regression for LSD and regression for SSD, respectively. Still, performance is not better then Meta-switch, even worser in the distal case. This is because, in CNN-switch, CNN instances for pose estimation has only seen one of the datasets, either LSD or SSD.  Whereas, in Meta-switch, feature encoder has seen both LSD and SSD dataset during the training, acquiring more robustness. 

Finally, in transition from proximal to distal case, performance remains almost same for both, Implicit-switch and Meta-switch. Whereas, for all the other methods, performance degrades by a large margin. This indicates the effectiveness of our proposed switching strategies in getting a more reliable and robust visual servo control employing state-of-the-art CNN for relative pose estimation. 

\section{CONCLUSIONS}
\label{sec:conclusion}
We have presented a CNN based method to perform eye-in-hand pose based visual servoing with a manipulator robot for static scenes. Our CNN architecture design achieves better pose regression accuracy than previous related works.  
It has been shown that the CNN switching method has been effective in improving the precision of visual servo output at the finer level in comparison to the naive method.
For efficient and roboust CNN switching, a meta learning approach called model-agnostic-meta-learning (MAML) is used to train a single model which is good for all three different tasks. 

The training and testing scenes taken into considerations in this work are somewhat simpler. In future extension of this work, more complex and practical environments could be included for evaluations. One of the challenges with CNN based methods for visual servoing is the need for real-time operation. In our work, camera motion control was saccadic due to low fps CNN operation (i.e. around 2-3 fps) on a desktop system without GPUs. Also, it would be interesting to have a future study on control theoretic guarantee for visual servo output.

\bibliographystyle{IEEEtran}
\bibliography{IEEEabrv,vs}

\end{document}